\documentclass[11pt,a4paper]{article}
\usepackage[hyperref]{emnlp-ijcnlp-2019}
\usepackage{times}
\usepackage{latexsym}

\usepackage{graphicx}
\usepackage[labelformat=simple]{subcaption}
\usepackage{url}

\aclfinalcopy % Uncomment this line for the final submission

\title{Experiments with LVT and FRE for Transformer model}

\author{
  Ilshat Gibadullin \\
  {\tt i.gibadullin@innopolis.ru} \\\And
  Aidar Valeev \\
  {\tt ai.valeev@innopolis.ru} \\}

\date{}

\begin{document}
\maketitle
\begin{abstract}
In this paper we experiment with Large Vocabulary Trick and Feature-rich encoding applied to Transformer model for Text Summarization. We could not achieve better results, than the analogous RNN-based sequence-to-sequence model, so we tried more models to find out, what improves the results and what deteriorates them.
\end{abstract}

\section{Introduction}

There are quite a lot of researches of additional busting for RNN-based Sequence-to-sequence (seq2seq) models, but due-to the novelty, there is a lack of them for Transformer model. Transformer model is becoming the state-of-the-art in machine translation task, showing significant improvements over the seq2seq models, but the performance of Transformer model in abstract-summarization task is not explicitly investigated. That motivated us to try to apply a couple of architecture independent methods to Transformer model: Large Vocabulary Trick and Feature-rich encoding, which gave improvements with seq2seq models, and compare the obtained results with the base Transformer model results and the ones obtained and evaluated in \newcite{nallapati}.

The paper is laid out as follows: Section 2 gives a short review of related work;  Section 3 describes the approaches we applied; Section 4 describes the data and the system; Section 5 reports the results and their analysis; and finally Section 6 sums it all up.

\section{Related Work}

Currently there are several fundamental Neural Machine Translation models competing to be the state-of-the-art, which are also applicable for Text Summarization: recurrent and convolutional neural networks \cite{bahdanau2015s2s,gehring2017conv}, and more recent attention-based Transformer model \cite{vaswani2017transformer}. All models consist of encoder-decoder parts, but in the first one both encoder and decoder are LSTM layers with an attention layer between them, second one is based on convolutions, while in Transformer encoder stack consists of a number of layers composed of a multi-head self-attention layer and a feed-forward neural network, and decoder stack --- same but each layer has an intermediate attention layer with encoder stack input. Default Transformer model uses Byte-Pair Encoding (BPE) \cite{bpe} to encode the text data.

Motivated by \cite{nallapati}, we decided to apply Large Vocabulary Trick (LVT) \cite{lvt} and Feature-rich encoding (FRE) to Transformer model. The main idea of LVT is to only work with a subset of the vocabulary, which is relevant to the current processing batch - the words from the batch and most frequent words to fill up till the limit. LVT allows to considerably lower the training time when the vocabulary is very large. FRE encodes additional information to the input - for each word there are parts-of-speech (POS) and named-entity tags (NER), term frequency (TF) and inverse document frequency (IDF) statistics.

\section{Models}

\subsection{BPE-based Transformer model}

For comparison purposes, we used Transformer with default settings, as was described in \cite{vaswani2017transformer}. The Byte-Pair Encoding (BPE) \cite{bpe} was used to encode the input sequence, where the size of sub-words vocabulary was set to 32K. The embedding layer was shared between the encoder and decoder parts of the model and was initialized randomly.

\subsection{Baseline}

As a baseline model we took Transformer model based on words vocabulary instead of BPE. The size of the vocabulary obtained from dataset is 123K. The embedding layer was separated for the encoder and decoder parts of the model. Initialization of the weights for these layers was also performed randomly.

\subsection{FRE fit-to-hidden (fre-f2h)}

This model uses Feature-reach encoding technique to extend the words embedding vector of Transformer encoder part's input, so the output of encoder embedding layer is the vector, which is a concatenation of the following sub-vectors: the word embedding vector, POS of the word, TF and IDF of the word. POS vectors are represented by one-hot encoding. Continuous features ,such as TF and IDF, were converted into categorical values by discretizing them into a fixed number of bins, and one-hot encode to indicate the bin number they fall into. Word embedding weights were initialized randomly. The embedding layer of the decoder part of Transformer was the same as in Baseline model described in the previous subsection. The vectors obtained by the encoder and decoder embedding layers should have dimensions equal to the hidden size of Transformer model. Thus, we should limit the dimension of the sub-vectors dimensions in the encoder embedding layer, so the dimension of obtained vectors should fit the hidden size of the model.

\subsection{FRE linear-map-to-hidden (fre-lm2h)}

The difference from previous described model is that here we use an additional linear layer without bias in the encoder embedding layer to map the vector obtained by the concatenation of sub-vectors to hidden size of Transformer model. Thus, the dimension of the vector obtained by concatenation of sub-vectors does not have to fit the hidden size of the model.

\subsection{FRE + LVT}

In this model we change the embedding layer of decoder part of Transformer model by Large Vocabulary Trick approach. For each training batch, we build new batch vocabulary by words from all texts in this batch. Required decoder vocabulary size is set to 2K, so in case of a lack of words in vocabulary obtained from the batch texts, we extend it by the most frequent words. The weights of decoder embedding layer is the same as in previous models, but during training we use and modify only weights of those words, which are in the current batch vocabulary. During inference we use whole vocabulary. The encoder embedding layer is the same as in the previous model.

\section{Experimental Setup}
\subsection{Data}
We used Gigaword corpus\footnote{https://github.com/alesee/abstractive-text-summarization} for training the models, it consists of 3.6 million article-title pairs. We could not acquire the annotated version of it, so we annotated it by ourselves, but we did not add named-entity tags, because Named-entity Recognition tools we tried (StanfordNERTagger, SennaNERTagger) performed poorly, since the corpus was lower-cased and it was the only version we could find. We deduplicate and divide the data into 2 parts: validation set of 2000 sentences and the training one. Validation files were used to monitor the convergence of the training. 

We used DUC 2003 corpus for testing the models, so that we could also compare our results with the results in the paper \cite{nallapati}.

\subsection{System Setup}
Transformer model \cite{vaswani2017transformer} with base setting from tensorflow/official/models\footnote{https://github.com/tensorflow/models} was used in the experiments. To evaluate the quality of the summarization, Recall-Oriented Understudy for Gisting Evaluation (ROUGE) metric \cite{rouge} was used.

\subsection{Hardware}
Since most of the operations inside the model were numeric and easily parallelizable, NVIDIA GTX 1080 Ti with GPU memory 11 GB was used to speed up the process.

\section{Results}
Firstly, we trained the BPE-based Transformer model for 5 epochs, each epoch took 3 hours 3 minutes. We got good results: 7.96 Rouge-2 and 21.54 Rouge-L (Table \ref{duc_test}).

Secondly, we trained baseline model also for 5 epochs, each one took 3 hours 52 minutes. The results got worse: 7.31 Rouge-2 and 20.29 Rouge-L (Table \ref{duc_test}).

Thirdly, FRE fit-to-hidden was trained also for 5 epochs, each took 3:54. The results got worse again: 6.06 Rouge-2 and 18.88 Rouge-L (Table \ref{duc_test}).

\begin{figure}
\centering
\includegraphics[width=1\linewidth]{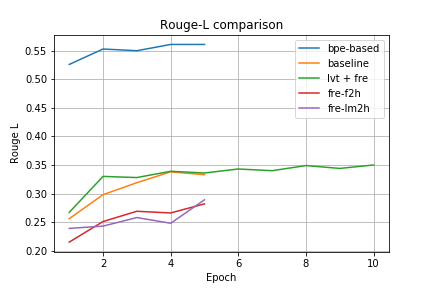}
\caption{Rouge-L validation scores.}
\label{fig:rouge_l_comp}
\end{figure}

Fourthly, we tried FRE linear-map-to-hidden also for 5 epochs, each epoch took 3:55. The results on DUC 2003 got worse again: 5.87 Rouge-2 and 18.34 Rouge-L (Table \ref{duc_test}), but validation scores with fre-f2h are very close and Rouge-L outperformed it on the 5th epoch, as can be seen in Figure \ref{fig:rouge_l_comp}.

\begin{figure}
\centering
\includegraphics[width=1\linewidth]{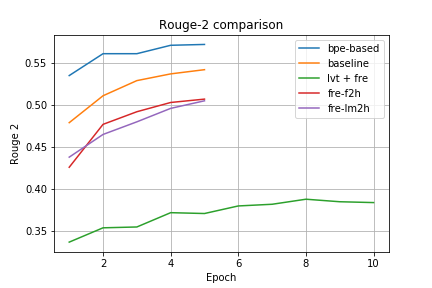}
\caption{Rouge-2 validation scores.}
\label{fig:rouge2_comp}
\end{figure}

\begin{table}
\begin{center}
\begin{tabular}{|l|r|r|r|}
\hline \bf Model & \bf Rouge1 & \bf Rouge2 & \bf RougeL \\ \hline
bpe-based & 23.43 & 7.96 & 21.54 \\ \hline
baseline & 21.98 & 7.31 & 20.29 \\ \hline
fre-f2h & 20.53 & 6.06 & 18.88 \\ \hline
fre-lm2h & 20.01 & 5.87 & 18.34 \\ \hline
fre+lvt & 9.62 & 1.47 & 9.16 \\ \hline \hline
TOPIARY & 25.12 & 6.46 & 20.12 \\ \hline
ABS & 26.55 & 7.06 & 22.05 \\ \hline
ABS+ & 28.18 & 8.49 & 23.81 \\ \hline
RAS-Elman & 28.97 & 8.26 & 24.06 \\ \hline
words-lvt2k-1s & 28.35 & 9.46 & 24.59 \\ \hline
words-lvt5k-1s & 28.61 & 9.42 & 25.24 \\ \hline
\end{tabular}
\end{center}
\caption{\label{duc_test} DUC 2003 test scores. There are our models' scores in the upper part and the scores from the paper \cite{nallapati} in the lower part.}
\end{table}

\begin{figure}
\centering
\includegraphics[width=1\linewidth]{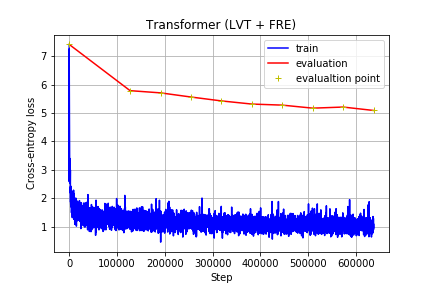}
\caption{Cross-entropy loss during training of FRE+LVT model.}
\label{fig:lvtfre_loss}
\end{figure}

\begin{figure}
\centering
\includegraphics[width=1\linewidth]{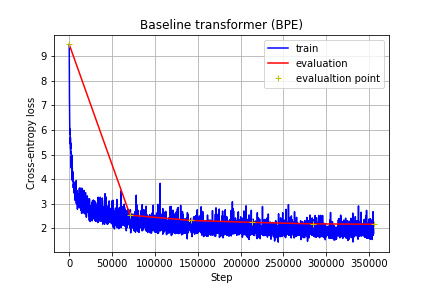}
\caption{Cross-entropy loss during training of BPE-based Transformer model.}
\label{fig:bpe_loss}
\end{figure}

Finally, we trained LVT+FRE for 10 epochs, each took 2:18. The results plummeted: 1.47 Rouge-2 and 9.16 Rouge-L (Table \ref{duc_test}). Most probably, 10 epochs were not enough for embeddings to train, since only 2K of them were updating each batch. In the Figure \ref{fig:lvtfre_loss}, we can see that the cross-entropy loss (blue) of LVT+FRE is approximately the same, as in the Figure \ref{fig:bpe_loss} with BPE-based model, but the evaluation line (red) is much higher, meaning worse. This is because the latter is computed using the whole vocabulary, while the former - using 2K vocabulary, which is unique for each batch.

\section{Conclusion}

In this work we evaluated default BPE-based Transformer model, Transformer with words vocabulary as baseline and tried to apply FRE and LVT approaches over it. Validation scores showed that these approaches do not give improvements and even worsen the quality against baseline. We found out that the default BPE-based Transformer model gives the best result among all evaluated models. We used DUC 2003 dataset as a test set to then compare our models with the models evaluated in \cite{nallapati}. BPE-based Transformer model outperforms TOPIARY and ABS models, but performs worse than the models proposed by authors of \cite{nallapati}. FRE and LVT approaches also performs worse than the baseline, while it performs worse than the BPE-based model.

We think that FRE doesn't give improvements over baseline, because the quality of Gigaword annotated by ourselves is worse than original annotated Gigaword dataset. The LVT requires more iterations to converge, but the convergence is too slow, so we not sure that it will increase even till baseline results. Thus, application of LVT to Transformer doesn't make any sense in the form in which we described it, because even if it shorten the training time of one epoch, it doesn't improve the overall result.

Also, we did not try any model with the pretrained word embeddings

\begin{table*}
\begin{center}
\begin{tabular}{|l|}
\hline 
\bf Source Document \\ \hline
schizophrenia patients whose medication could n't stop the imaginary voices in their \\ heads gained some relief after researchers repeatedly sent a magnetic field into a \\ small area of their brains . \\ \hline
\bf Ground truth Summary \\ \hline
Magnetic pulse series sent through brain may ease schizophrenic voices \\ \hline
\bf BPE-based \\ \hline
schizophrenia patients gain some relief \\ \hline
\bf Baseline \\ \hline
study shows link between schizophrenia patients \\ \hline
\bf FRE-f2h \\ \hline
study links schizophrenia to schizophrenia \\ \hline
\bf FRE-lm2h \\ \hline
researchers say they can t stop some people from schizophrenia \\ \hline
\bf LVT + FRE \\ \hline
nasal implants pose dilemma \\ \hline \hline

\bf Source Document \\ \hline
china was evacuating 330,000 people friday from land along the raging yangtze river \\
that officials were preparing to sacrifice to flooding to safeguard cities downstream . \\ \hline
\bf Ground truth Summary \\ \hline
Chinese military personnel conducting extensive flood control efforts along Yangtze. \\ \hline
\bf BPE-based \\ \hline
china orders soldiers to fight to the death \\ \hline
\bf Baseline \\ \hline
china orders soldiers to fight yangtze floods \\ \hline
\bf FRE-f2h \\ \hline
china orders soldiers to fight floods \\ \hline
\bf FRE-lm2h \\ \hline
china orders soldiers to fight floods \\ \hline
\bf LVT + FRE \\ \hline
china mobilizes soldiers to safeguard potable reservoirs \\ \hline \hline

\bf Source Document \\ \hline
the czech republic and hungary will not compete with each other in their bids to join \\ nato and the european union -lrb- eu -rrb- , the hungarian telegraph agency reported \\ today . \\ \hline
\bf Ground truth Summary \\ \hline
Czech Republic, Hungary vow not to compete for NATO bid. \\ \hline
\bf BPE-based \\ \hline
czech republic hungary not to compete in nato \\ \hline
\bf Baseline \\ \hline
czech hungary not to compete for nato eu membership \\ \hline
\bf FRE-f2h \\ \hline
czech republic hungary to compete in nato \\ \hline
\bf FRE-lm2h \\ \hline
hungary czech republic not to compete in nato bids \\ \hline
\bf LVT + FRE \\ \hline
prague czechs not intend to participate in joining nato \\ \hline
\end{tabular}
\end{center}
\caption{Examples.}
\end{table*}

\bibliography{nlp}
\bibliographystyle{acl_natbib}

\end{document}